\documentclass[submission,copyright,creativecommons]{eptcs}
\providecommand{\event}{SOS 2007} 

\usepackage{iftex}
\usepackage{amsmath}
\usepackage{amssymb}
\usepackage{todonotes}
\usepackage{makecell}
\usepackage{tcolorbox}
\usepackage{longtable}

\ifpdf
  \usepackage{underscore}         
  \usepackage[T1]{fontenc}        
\usepackage{latexsym}
\usepackage{tikz}
\usetikzlibrary{shapes.geometric, arrows}
\tikzstyle{startstop} = [rectangle, rounded corners, minimum width=3cm, minimum height=1cm,text centered, draw=black, fill=red!30]
\tikzstyle{process} = [rectangle, minimum width=3cm, minimum height=1cm, text centered, draw=black, fill=orange!30]
\tikzstyle{io} = [trapezium, trapezium left angle=70, trapezium right angle=110, minimum width=3cm, minimum height=1cm, text centered, draw=black, ill=blue!30]
\tikzstyle{decision} = [diamond, minimum width=3cm, minimum height=1cm, text centered, draw=black, fill=green!30]
\tikzstyle{arrow} = [thick,->,>=stealth]
\usetikzlibrary{positioning,arrows.meta}
\usepackage{changepage}

\usepackage{graphicx}
\else
  \usepackage{breakurl}           
\fi

\title{Using Large Language Models for (De-)Formalization and Natural Argumentation Exercises for Beginner's Students}
\author{Merlin Carl
\institute{EUF\\ Flensburg, Germany}
\institute{Institut f\"ur Mathematik\\
Europa-Universit\"at Flensburg\\
Flensburg, Germany}
\email{merlin.carl@uni-flensburg.de}
}
\def\titlerunning{Deformalization Exercises with Large Language Models}
\def\authorrunning{M. Carl}
\begin{document}
\maketitle

\begin{abstract}
We describe two systems currently being developed that use large language models for the automatized correction of (i) exercises in translating back and forth between natural language and the languages of propositional logic and first-order predicate logic and (ii) exercises in writing simple arguments in natural language in non-mathematical scenarios. 
\end{abstract}

\section{Using Large Language Models for Autoformalization}

Autoformalization, i.e., the automated translation from natural language to formal logic, is a natural language processing task that has been adressed in a number of ways; natural language proof checking systems such as Naproche (see, e.g., Cramer, \cite{Cr}) and SAD (Verchinine et al., \cite{VLP}) make use of grammar-based approaches, with natural-language parsers producing intermediate formats between natural language and formal logic. Recently, the interest in machine learning approaches has increased, including the use of neural networks (Azerbayev et al., \cite{APSARA}, Wang et al., \cite{WBKU}) and large language models (Wu et al., \cite{WJLRSJS}); autoformalization for mathematics at the undergraduate level was explored in Azerbayev et al. \cite{APSARA}. Two strong large language model that are publically available are OpenAI's GPT-3.5, with the associated text completion model text-davinci-003\footnote{See \url{https://platform.openai.com/docs/models/gpt-3-5}.}, and the more recent GPT-4-Turbo.\footnote{See \url{https://platform.openai.com/docs/models/gpt-4-and-gpt-4-turbo}.} One advantage of pretrained models over training neural networks is that frequently, a few examples are sufficient to obtain a stable and usable model. It has been observed in \cite{APSARA} that GPT can successfully be used for autoformalization. Not surprisingly, the performance improves drastically when one merely demands autoformalization of expressions in a controlled natural language (CNL) rather than arbitrary sentences in natural (mathematical) language, which can be used, e.g., for automated proof checking in teaching contexts, see \cite{C1}.

For the applications we describe in this article, however, we are interested not in translating mathematical sentences into formal logic, but rather statements in natural language on everyday matters. To this end, we wrote prompts for text-davinci-003 consisting of different numbers of examples, depending on the complexity of the task:

\begin{itemize}
\item 30 examples for the translation of natural language sentences into formulas in propositional logic with a given notation.
\item 44 examples for the translation of natural language sentences into formulas in propositional logic, combined with a classification determining whether that sentences formulates a claim or an assumption.
\item 56 examples for the translation of natural language sentences into formulas in first-order predicate logic.
\end{itemize}

Typical example prompts for the three tasks looked like this (original German; translated to English for the convenience of the reader):

\begin{itemize}
\item notation:\{S:Fritz takes a boat;F:Fritz takes a plane;A:Fritz arrives in America;K:Fritz tries to swim\}Fritz arrives in America if and only if he takes a boat or a plane, but not if he tries to swim.\footnote{Note that the system is to be used by students in Germany.}\#$((S\vee F)\leftrightarrow A)\wedge (K\rightarrow \neg A)$\S
\item notation:\{W:This is supposed to be a joke;L:This is supposed to be funny;N:This is new\}If this is supposed to be a joke, it is neither funny nor new.$\sharp [claim\footnote{Translated for the convenience of the reader; in the actual prompt, we used ``beh'' for ``Behauptung'', which represents claims in the internal Diproche format, see \cite{C}. For assumptions, it would have been ``vss'' for ``Voraussetzung''.},[W,\rightarrow,[neg,[L,or,N]]]]$\S
\item notation:\{B(x,y):x is the brother of y;S(x,y):x is the sister of y\}The sister of someone's brother is that someone's sister.$\sharp \forall{x}:\forall{z}:(\exists{y}:(B(x,y)\wedge S(y,z))\rightarrow S(x,z))$\S
\end{itemize}

As one can see, these examples are structured as follows: the first part of the form ``notation:\{...\}'' introduces a number of propositional letters, predicate letters and constant symbols, along with their intended semantics; this is followed by a sentence in natural language, a sharp serving as a separation symbol, a formalization of the natural language sentence in the given notation and a stop symbol. In each case, a few examples were added where either the natual language sentence was not a sentence at all, but rather some nonsense string, or could not be expressed in the given notation.

Requests to the model prompted in this way can then be made by expanding the prompt by a string of the form ``notation:\{...\}$\phi\sharp$'', where $\phi$ is a natural language sentence. Experiments showed a satisfying performance on the intended kind of (simple) example sentences. An impressive feature was that formalizations worked well even when the precise formulation did not use the expressions given in the prompt. Thus, in the notation ``$R$: It rains, $S$: There is a storm; $P$-the party will be cancelled'', the sentence ``If it pours or there is a strong wind, there will be no feast'' was (correctly) formalized as $(R\vee S)\rightarrow P$.

The main drawbacks were the following:

\begin{itemize}
\item The model showed a tendency to report sentences with strange, wrong or absurd content, such as ``If the moon is made of green cheese, then there is a giraffe on the moon'' are as erroneous, even though they could easily be formalized within the given notation. Adding an explanation of the specific meaning of ``error'' in this context and several further examples did not solve this issue. Consequently, such examples, which are somewhat typical for logic classes, should at the moment be avoided when using the model.
\item The model showed a certain tendency to use all pieces of the given notation in its formalization, so that, e.g., ``Barking dogs don't bark''\footnote{Note that this is not a typo. In order to evaluate how true the model formalizes the given sentence, rather than some more common substitute for it, we deliberately used this (absurd) sentence rather than the well-known proverb that barking dogs don't bite.} was formalized as $\forall{x}((D(x)\wedge B(x))\rightarrow S(x))$ in the notation where $D(x)$ stood for ``$x$ is a dog'', $B(x)$ for ``$x$ barks'' and $S(x)$ for ``$x$ bites''. This was considerably improved by adding several examples with superfluous notation.
\item The model showed a tendency to ``project'' expressions onto the given notation; for example, in the notation from the last bullet point, ``Barking cats don't bite'' was formalized as $\forall{x}((D(x)\wedge B(x))\rightarrow S(x))$, even though no predicate letter for ``cat'' was contained in the notation (``meowing cats don't bite'', however, led to an error message). Whether or not this presents a serious issue for the intended application remains to be seen.
\item For sentences of high logical complexity (e.g., containing several quantifier alternations or junctors), the formalizations were frequently wrong. This, however, does not present much of an issue for the intended application.
\item The disadvantages of using cloud-based LLMs include (i) their lack of stability (continued training can lead to a much worse performance, rendering working applications unusable), (ii) their lack of reliable availability (models may stop being available altogether, and temporarily become unavailable due to server capacity issues) and (iii) pricing. For these reasons, we expect our system to remain in an experimental state until workable local alternatives become available.

\end{itemize}

By now, text-davinci-003, which was state of the art when our system was developed early in 2023, is considered ``legacy'' by OpenAI and its use deprecated. Thus, additionally, we tried the same task with an AI-``assistant'' based on the more recent model GPT-4-Turbo. Here, we achieved a surprisingly good performance using merely a prompt explaining the notation to be used (i.e., without offering any initial example cases); in contrast to the above examples, however, the notation -- i.e., the abbreviations for the atomic propositions, predicate and constant symbols to be used in the formalization -- was fixed and the same for all the examples. The precise prompts, along with the example sentences and the obtained output for $50$ instances of first-order-formalization and $57$ instances of formalization in propositional logic, can be found in the appendix. As can be seen from the examples (54)-(57) for propositional logic, ``nonsense'' sentences concerning strongly counterfactual scenarious did no longer pose an issue for GPT-4-Turbo. Concerning logically absurd sentences, such as (43), (44), (47), (48), one of these (44) was still (wrongly) labeled as ``not expressable'',\footnote{When asked for the reason why (44) would not be expressable, the model replied that ``the statement is self-contradictory and cannot be expressed meaningfully using the given notation or any standard logical operators because it violates the law of non-contradiction. Thus, the proper response is ``not expressable''.'', thus explicitly defending the (wrong) claim that logically inconsistent statements cannot be expressed in the formalism of propositional logic.} while the other two were processed correctly. In formalization in quantifier logic, all ``absurd'' examples, namely (24), (26), (33), were processed correctly. While a certain flexibility of expression was retained -- for example, the sentence ``In terms of size, Fritz surpasses himself'' was formalized correctly given a notation that contained a two-place predicate for ``larger than'' -- the issue of replacing non-expressible terms by terms in the vocabulary did no longer show up: Thus, the propositional examples (20), (21), (27), (57) were correctly identified as ``not expressable''.

For the task of formalization in propositional logic, we also evaluated several large language models that are available locally, i.e., they can be run on the user's local machine rather than remotely. The performance of general local LLMs and also of LLMs focusing on mathematics -- such as WizardMath-70B\footnote{See \url{https://huggingface.co/TheBloke/WizardMath-70B-V1.0-GGML}.} -- turned out to be poor; the only reasonable results were obtained with models trained on the task of code-writing, in particular WizardCoder-34B.\footnote{See \url{https://huggingface.co/WizardLM/WizardCoder-Python-34B-V1.0}.}

For propositional logic, GPT-4-Turbo formalized $55$ of the $57$ sample sentences correctly, or about 96.5 percent. The best performance of a local large language model for the same set of sentences was successful for $40$ sentences, corresponding to a success rate of about 69 percent. As can be seen in lines (54)-(57), highly counterfactual content -- which was a serious issue for davinci-003 -- was consistently processed correctly by both models. Logically absurd statements, such as (43), (44), (48) were falsely labeled as ``not expressible'' once by both models; here, GPT-4 generated a lengthy explanation claiming that contradictory statements cannot be expressed in formal logic. Tautologous statements, such as (47) were unproblematic, as were factually wrong statements for both models. Both models could successfully handle a certain freedom of expression, as is demonstrated by the sentences (17), (18), (19), (31), (32), (34). Both models still had an issue with intricate logical structure, such as triple negation (38); as such examples are likely to arise rarely in the training data, this is to be expected. While GPT-4-Turbo clearly outperforms the local model, both models show usable results for the intended application, namely simple exercises of limited logical complexity. 

For the task of autoformalizing natural language sentences in first-order logic, no local LLM showed a satisfying performance. The best results were obtained with GPT-4-Turbo, where $46$ out of $50$ provided example sentences were formalized correctly, i.e., $92$ percent. The unexpressable sentences (18), (32), (36) were correctly identified. In cases where the content was counterfactual or absurd, the model stated this fact in a text comment accompanying the formalization, but still provided a correct formalization (see (24), (26), (27), (33)). For some of those sentences that were not formalized as expected, the model provided explanations: In the case of (13) (``If Hector barks, he is not a real dog''), the model explained that the notation did not allow for expressing ``is not a real dog''; given common ways of using ``real XYZ'', this is arguably correct; a similar point can be made for (46). Without these examples, the success rate would be about 96 percent. An actual mistake is (34), which was not expressible, as no 2-place-predicate ``$x$ bites $y$'' was given in the notation; in (35), the formulation ``baying canine'' was apparently too ``off'' in order to be identified as ``barking dog''. However, such formulations are not to be expected from someone seriously attempting to solve a deformalization exercise, so that the practical relevance of this mistake for the intended application appear limited.

We point out that the system is currently in an experimental stage. In particular, several practical challenges in terms of pricing and stability of the underlying LLMs will need to be overcome before the system can be actually employed (and tested) in teaching.

\section{Automatized Correction of (De-)Formalization Exercises}

Beginner students frequently have difficulties both with expressing statements in the language of logic and with interpreting statements written in the language of formal logic. Even after  becoming acquainted to the meaning of the logical symbols, extracting the meaning remains a challenge: Frequently, a sentence such as $\exists{x}:\text{child}(x)\wedge\text{swims}(x)$ will be read out as ``There is $x$ such that $x$ is a child and $x$ swims'', remaining close to the surface structure of the formula, rather than something like ``Some child swims''.

Indeed, the relation between ``natural'' ways of thinking of something and the way it is expressed in formal logic is quite intricate; one notorious example of this being the way of expressing dynamics via quantifier changes.\footnote{For example, in the definition of continuity, the idea of ``moving one point closer to another'' is a dynamical conception expressed via quantifier changes; for a detailed discussion, see the considerations of the concept of continuity in Lakoff and Nunez \cite{LN}, p. 309-315.}

Formalization and deformalization exercises are meant to practice this crucial skill in university education in mathematics. In a formalization exercise -- also known as ``math dictation'', see, e.g., \cite{C2}\footnote{We originally learned this term from Michael Junk.} -- a sentence in natural language is given, together with some formal vocabulary, and the student's task is to produce a logical formula expressing this sentence. Thus, a typical formalization exercise could look like this:

\begin{itemize}
\item Let $S$ stand for the statement ``The sun shines'', $W$ for the statement ``I go out for a walk''. Formalize the statement ``I won't go for a walk unless the sun shines'' in the language of propositional logic.
\item Let $C(x)$ stand for ``$x$ is a child'', $S(x)$ for ``$x$ swims''. Formalize the statement ``Some children swim'' in the language of first-order predicate logic.
\end{itemize}

It is not too hard to provide automated feedback for such formalization exercises: One can, for example, store the exercise in the form of a pair $(\eta,\phi)$ consisting of the natural language statement $\eta$ to be displayed to the user and a correct formalization $\phi$ of it, take the user's input, determine whether it is a well-formed formula at all (and, if not, provide according feedback), and, in case it is, pass the tasks $\phi\rightarrow\eta$ and $\eta\rightarrow\phi$ to an automated theorem prover, such as the PyProver, which can be imported into Python code as a package. Systems that work roughly in this way include the ``mathematical logic tutor'' of Moreno et al. \cite{BM} or the formalization exercises in Edukera \cite{Edukera}.

We remark here\footnote{We thank one of our referees for pointing out that this point warrants a discussion.} that the concept of formalization is rather intricate. Indeed, a formalization in the sense above is a special case of a translation, and so the difficult question about the adequacy of translations also applies to formalizations: When is a formula $\phi$ an adequate formalization of a natural language sentence $S$? A naive approach might be to say that $\phi$ and $S$ must be in some sense ``provably equivalent''; this, however, leads to several difficulties: First, one would have to make precise the informal notion of proof required to make sense of this criterion. Second, such a notion would necessarily be relative to the set of accepted background assumptions, which are usually left implicit and may be highly context-dependent. For example, when asked to formalize ``If $a$, $b$ are the lengths of the catheti of a right triangle and $c$ is the length of its hypotenuse, then $c^3>a^3+b^3$'', it should be acceptable -- if theory-laden -- to respond with $\forall{a,b,c\in\mathbb{R}^{+}}((c>a+b\wedge a^2+b^2=c^2)\rightarrow c^3>a^3+b^3)$, thus presupposing Pythagoras' theorem; but if the sentence would be to formalize ``If $a$, $b$ are the lengths of the catheti of a right triangle and $c$ is the length of its hypotenuse, then $a^2+b^2=c^2$'', it would be most inadequate to formalize this in the same way, thus leading to a tautology. (One might even ask whether it can be formalized adequately in the language of real closed fields at all without circularity.) Third, mere provable equivalence is a poor criterion for the adequacy of a formalization; for example, formalizing either of the preceding statements as $1=1$ should certainly not be counted as correct.\footnote{Note that the former statement is correct, as $a^3+b^3<a^2c+b^2c=c^2\cdot c=c^3$.} Other great examples of formalization, such as Gödel's arithmetization of the concept of first-order provability require a creative mastering of some background theory. This kind of formalizing is a task resembling programming more than translation, and it is not what the system sketched in this paper is meant to teach. The difference between ``translation-like formalization'' and ``programming-like formalization'', although didactically quite obvious, is not easy to pin down logically. However, by considering everyday rather than mathematical contexts, this issue is mostly avoided: There is not much background theory that could be used in expressing a sentence such as ``Barking dogs don't bite'' into a first-order language with a vocabulary for ``barks'', ``bites'' and ``is a dog'', let alone for interpreting the respective first-order formula in natural language. We will thus accept a formula $\phi$ as a correct formalization of a sentence $S$ with a fixed given formalization $\psi$ if and only if the equivalence $\phi\leftrightarrow\psi$ is a tautology in the respective logic (that is, propositional logic or first-order logic), i.e., using the empty background theory. This seems to yield a didactically acceptable notion of adequacy, as long as the sentences under consideration are neither tautological nor contradictory -- although ``weird'' formalizations accepted by a system working with this premise are still possible (e.g. by writing something like $A\vee (A\wedge A)$ or $\neg\neg\neg\neg A$ rather than $A$, or by forming a conjunction with a number of unrelated tautologies), they are unlikely to actually be proposed by students, except by those who have already mastered the subject at hand anyway.

Deformalization exercises, on the other hand, are a far more delicate matter. In a deformalization exercise, a student is given a formal vocabulary together with a logical formula using that vocabulary, and is asked to express it in natural language in the simplest possible terms. Thus, typical deformalization exercises look as follows:

\begin{itemize}
\item Let $S$ stand for the statement ``The sun shines'', $W$ for the statement ``I go out for a walk''. Express the statement $W\rightarrow S$ in natural language.
\item Let $C(x)$ stand for ``$x$ is a child'', $S(x)$ for ``$x$ swims''. Express the statement $\exists{x}:C(x)\wedge S(x)$ in natural language.
\end{itemize}

The automatized correction of such exercises is challenging in at least to ways: First, one needs to automatically translate the user's natural language input into the appropriate formal language, using the specified vocabulary. Second, one needs to grade the ``naturalness'' of the user's input, so that expressions such as ``There is $x$ such that $x$ is a child and $x$ swims'' mentioned above receive a different feedback than ``Some children swim''. 

In our system, which is currently in an experimental stage but will eventually be integrated into the Diproche system (\cite{CK}, \cite{C}, \cite{C1}, \cite{C2}, \cite{C3}), exercises are stored as triples $(n,\eta,\phi)$, where $n$ is the available vocabulary, $\eta$ is a natural language sentence and $\phi$ is the formalization thereof, which should be ``optimal'' in terms of naturalness. This kind of representation has the advantage that the same exercise can both be used as a formalization exercise (displaying $n$ and $\eta$, then asking for $\phi$) and as a deformalization exercise (displaying $n$ and $\phi$, then asking for $\eta$). 

\subsection{Checking Deformalizations}

In this section, we will explain the architecture of the checking routine for deformalization exercises. As explained above, users are given a set of notations and a formula using these notations. They can then enter an arbitrary string in a text window in a web interface. This input string is passed on to the formalization routine, which uses the input and the training examples to generate a request for a large language model (such as text-davinci-003) and passes it on to this LLM; if this yields the output ``error'', an error message is displayed to the user and no further processing takes place; otherwise, the formalization of the users input and the formula given in the problem statement are passed on to the PyProver, which determines whether the implications between them are provable (i.e., propositional or first-order tautologies). More precisely, if $\psi$ is the formal expression fixed in the problem statement and $\phi$ is the result of autoformalizing the user's input, the PyProver is asked to prove both $\phi\rightarrow\psi$ and $\psi\rightarrow\phi$. The result of this is then passed on to the feedback creation, which will report which of these implications could be verified. 

However, a merely logically correct answer does not imply that the input string is a good deformalization of the given formula, or that the user has understood this formula. Typically, beginners will translate a formula such as $\forall{x}((D(x)\wedge B(x))\rightarrow \neg S(x))$ -- in the vocabulary given above -- ``word for word'' into something like ``For all $x$, if $x$ is a dog and $x$ barks, then $x$ does not bite''. Thus, in order to provide a meaningful assessment of the input, the mere logical correctness needs to be supplemented with an evaluation of the ``naturalness'' or ``simplicity'' of the input string. This is the task of the ``Grader'' module.

Our attempts to train language models to provide such an assessment were utterly unsuccessful so far: Even with a considerable number of examples, the model appeared to be unable to distinguish unnaturally formulated sentences from perfectly naturally formulated sentences with an unusual content. We thus resorted to a rather simple-minded solution, which, however, turned out to work quite well for our purpose: We measure the complexity of the input by relating its length to the length of the template solution entered as part of the problem statement (although not displayed to the user) and normalize the result. A ``word for word translation'' of logical syntax into natural language will usually be considerably longer than the shortest formulation in natural language, so that this can be expected to approximate the degree of naturalness satisfyingly well. More precisely, if $|s|$ denotes the length of a string, $\eta$ denotes the template solution in the problem statement and $\phi$ is the user's input, the degree of simplicity is given by

$$10\sigma(10(\frac{|\eta|}{|\phi|}-0.7))$$

where $\sigma$ is the usual sigmoid function, given by $\sigma(x):=\frac{1}{1+e^{-x}}$. This yields a measure of the input's simplicity on a scale from $0$ (not natural at all) to $10$ (very natural); feedback is provided depending on whether this value is less than or equal to $5$, strictly between $5$ and $8$ or equal to or above $8$. Thus, a good solution, which will have $|\eta|\approx|\phi|$, will have $\sigma(10(\frac{|\eta|}{|\phi|}-0.7))$ close to (but below) $1$, while a overcomplicated solution with $|\phi|$ much larger than $\eta$ will lead to a value close to $0$. A score of at least $5$ is awarded if the template solution has a length that is $70$ percent (or less) of the length of the input string.

Finally, the feedback creation either reports that the input is logically incorrect (not equivalent to the formula to be expressed) and, in this case, whether it is necessary, but not sufficient, sufficient, but not necessary, or neither; in this case, no evaluation of the simplicity of the input is given. If, on the other hand, the input was logically correct, then the user is provided with feedback reporting this, accompanied by the system's evaluation of its simplicity, asking the user, if necessary, to try her hand at further simplifications of her expression.

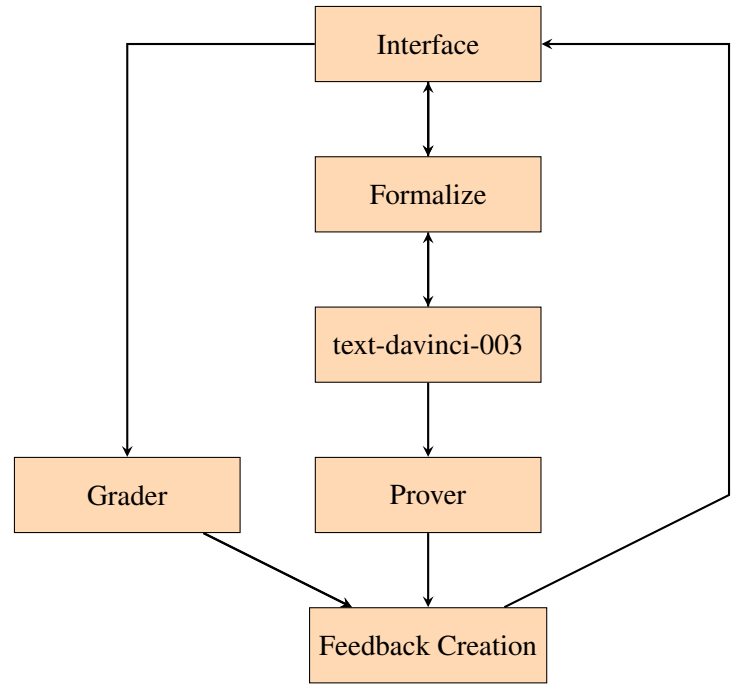
\begin{figure}[h]
\centering
\begin{tikzpicture}[node distance=2cm]

\node (interface) [process] {Interface};
\node (formalize) [process, below of=interface] {Formalize};
\node (textdavinci) [process, below of=formalize] {text-davinci-003};
\node (prover) [process, below of=textdavinci] {Prover};
\node (grader) [process, left of=prover, xshift=-2cm] {Grader};
\node (feedback) [process, below of=prover] {Feedback Creation};

\draw [arrow] (interface) -- (formalize);
\draw [arrow] (formalize) -- (textdavinci);
\draw[arrow] (textdavinci) -- (formalize);
\draw [arrow] (textdavinci) -- (prover);
\draw [arrow] (grader) -- (feedback);
\draw [arrow] (feedback) -- (4,-6) -- (4,0) -- (interface);

\draw [arrow] (grader) -- (feedback);
\draw [arrow] (prover) -- (feedback);

\draw [arrow] (interface) -- (-4,0) -- (grader);
\draw [arrow] (interface) -- (formalize);
\draw [arrow] (formalize) -- (interface);

\end{tikzpicture}
\caption{Flowchart for the correction routine for deformalization exercises. (Generated with the help of GPT-3.5.)}
\label{fig:grading}
\end{figure}

There are at least two sources of potential mistakes in this approach: First, the LLM may provide a formalization not faithful to the user's input, either by an actual formalization mistake or by interpreting the natural language semantics in a subtly different way, and second, the ATP may fail to verify the equivalence even if the input is correct. Due to the decidability of propositional logic (and the fact that the expressions arising in the given contexts will use less than $10$ propositional variables and be of surveyable length, so that no resource issues can occur), the latter kind of mistake can only occur for first-order logic. Since the formulas coming up in this context will be logically rather simple -- typically restricted to a single quantifier change -- this is not likely to be a frequent issue; however, a substantial evaluation of this point will have to wait until the system can be actually employed with student users.\footnote{One way to exclude such difficulties altogether would be to restrict exercises to decidable fragments of first-order logic, such as monadic first-order logic (see, e.g., \cite{B}), formulas with at most two variables (see, e.g., \cite{H}) etc.; these classes already lead to a rich supply of exercises. If this path is taken, one needs to ensure that the user's input will also fall into the relevant class.} Potentially, it cannot lead to wrong solutions being reported as correct, but it could lead to correct solutions being reported as incorrect. In order not to confuse users, the feedback should thus be formulated as a ``failure to verify'' rather than as a claim that the solution is wrong. The user may then attempt to reformulate her solution to make it easier processable. Another option would be to switch to an ATP that can generate countermodels to non-verifable statements and report these back to the user. The former type of mistake can lead both to correct solutions being reported as incorrect and to incorrect solutions being reported as correct. The latter issue may in particular come up for sentences that deviate from, but strongly resemble, very common natural language sentences that are likely to occur in the training data. Also, issues are likely to occur when students write ungrammatical inputs. Again, an evaluation of the seriousness of such issues will have to wait until the system can actually be employed. To counter such difficulties, the formalization obtained by the system could be reported back to the user to check whether it faithfully captures what she intended to express. However, since the system is intended to teach basic skills in translating back and forth between formal logic and natural language, it is not clear that the intended users will be able to determine whether the formalization was done correctly.

\section{Natural Language Argumentation}

Besides formalization, learning how to prove is another challenge for beginner students. The Diproche system (\cite{CK}, \cite{C}) is designed to provide automated feedback on the several aspects -- including logical correctness -- for solutions to simple beginner proving exercises written in a controlled natural language (CNL). Recently, large language models have been integrated in the Diproche architecture to provide a more ``liberal'' CNL than the original parser-based one (see \cite{C1}).

Diproche is focused on mathematical argumentation in areas such as propositional logic, Boolean set theory or elementary number theory. As a preliminary exercise, it may also be helpful to practice logical argumentation in non-mathematical contexts, putting aside the difficulties students may have with the mathematical content and symbolism. Thus, a natural argumentation exercise might look like this:

\begin{itemize}
\item Suppose that the following is true: If the sun shines, Hans goes for a walk. When Hans goes for a walk, he takes his dog with him. When Hans takes his dog for a walk, the dog barks at the cat on the neighbour's roof. When the dog barks at the cat on the roof, the cat runs away. However, the cat still sits on the roof. Show that the sun does not shine.
\end{itemize}

Using the same prompts discussed in the last section, such statements can automatically be translated into propositional logic\footnote{The same can, of course, easily be done with first-order logic. }; the resulting formal representation can then be passed on to the checking components of the Diproche system, which will provide feedback on the logical correctness of the argument. Thus, an accepted solution to the above exercise could look like this:

\begin{itemize}
\item The cat still sits on the roof. Hence the dog did not bark. Consequently, Hans did not take his dog for a walk. So Hans did not go for a walk. Thus the sun does not shine.
\end{itemize}

For mathematical exercises, the Diproche system gives feedback on linguistical correctness, logical correctness, type mistakes (using variables without introducing them before, or in a wrong way, such as adding two propositions), success in achieving proof obligations and supposed logical or algebraic fallacies. For the ``natural language argumentation'' exercises, the feedback on type mistakes will be suppressed, since this should not be an issue for the intended kind of argumentation. 

\section{Further Work}

Clearly, a system for didactical uses should be employed in teaching and evaluated in terms of usability and effectiveness. We plan to do so in the near future. On the technical side, the autoformalization with text-davinci-003 or GPT-4-Turbo, although impressive in many respects, still leaves some things to be desired.\footnote{Recall in particular the issues with relying on remote services rather than local models.} We plan to gather a substantial amount of training data and use it to fine-tune pretrained language models. Whether this will lead to improved performance remains to be seen. Concerning the \textit{formalization} exercises (``math dictations''), the LLM technology could be used for automatically deformalizing the user's formal input and reporting it back to her in the case the formalization is not correct, thus making it clearer where the mistake lies; here, more traditional NLP techniques, such as the ``pretty printer'' described in \cite{GGNS}, may also be useful.\footnote{We thank one of our anonymous referees for pointing out this reference to us.}

Meanwhile, we believe that good use could be made of the ``natural argumentation'' framework in other subjects, in particular in philosophy: Here, it is a basic type of exercise to reconstruct plain text arguments in a semi-formal style where assumptions and consequences are explicitly labeled, and the approach discussed here for the verification of natural language argumentation could then be used to verify whether an argument written in this way is in fact logically cogent.

\section{Acknowledgements}

We thank our three anonymous referees for several comments that helped in improving the presentation of the paper, along with constructive criticism concerning its content.

\bibliographystyle{eptcs}
\bibliography{deformalizationgpt}

\begin{thebibliography}{10}
\providecommand{\bibitemdeclare}[2]{}
\providecommand{\surnamestart}{}
\providecommand{\surnameend}{}
\providecommand{\urlprefix}{Available at }
\providecommand{\url}[1]{\texttt{#1}}
\providecommand{\href}[2]{\texttt{#2}}
\providecommand{\urlalt}[2]{\href{#1}{#2}}
\providecommand{\doi}[1]{doi:\urlalt{https://doi.org/#1}{#1}}
\providecommand{\eprint}[1]{arXiv:\urlalt{https://arxiv.org/abs/#1}{#1}}
\providecommand{\bibinfo}[2]{#2}

\bibitemdeclare{misc}{Edukera}
\bibitem{Edukera}
\emph{\bibinfo{title}{Edukera Homepage}}.
\newblock \bibinfo{note}{\url{https://www.edukera.com/}}.

\bibitemdeclare{article}{APSARA}
\bibitem{APSARA}
\bibinfo{author}{Zhangir \surnamestart Azerbayev\surnameend},
  \bibinfo{author}{Bartosz \surnamestart Piotrowski\surnameend},
  \bibinfo{author}{Hailey \surnamestart Schoelkopf\surnameend},
  \bibinfo{author}{Edward~W. \surnamestart Ayers\surnameend},
  \bibinfo{author}{Dragomir~R. \surnamestart Radev\surnameend} \&
  \bibinfo{author}{Jeremy \surnamestart Avigad\surnameend}
  (\bibinfo{year}{2023}): \emph{\bibinfo{title}{ProofNet: Autoformalizing and
  Formally Proving Undergraduate-Level Mathematics}}.
\newblock {\slshape \bibinfo{journal}{ArXiv}}, \doi{10.48550/arXiv.2302.12433}.
\newblock \eprint{arXiv:2302.12433v1}.

\bibitemdeclare{article}{B}
\bibitem{B}
\bibinfo{author}{Heinrich \surnamestart Behmann\surnameend}
  (\bibinfo{year}{1922}): \emph{\bibinfo{title}{Beitr\"age zur Algebra der
  Logik, insbesondere zum Entscheidungsproblem}}.
\newblock {\slshape \bibinfo{journal}{Mathematische Annalen}}
  \bibinfo{volume}{86}, pp. \bibinfo{pages}{163--229},
  \doi{10.1007/BF01457985}.

\bibitemdeclare{misc}{C2}
\bibitem{C2}
\bibinfo{author}{Merlin \surnamestart Carl\surnameend} (\bibinfo{year}{2020}):
  \emph{\bibinfo{title}{Automatized Evaluatoin of Formalization Exercises in
  Mathematics}}, \doi{10.48550/arXiv.2303.17513}.
\newblock \bibinfo{note}{ArXiv:2006.01800v2}.

\bibitemdeclare{article}{C}
\bibitem{C}
\bibinfo{author}{Merlin \surnamestart Carl\surnameend} (\bibinfo{year}{2020}):
  \emph{\bibinfo{title}{Number Theory and Axiomatic Geometry in the Diproche
  System}}.
\newblock {\slshape \bibinfo{journal}{Electronic Proceedings in Theoretical
  Computer Science}} \bibinfo{volume}{328}, pp. \bibinfo{pages}{56--78},
  \doi{10.4204/EPTCS.328.4}.

\bibitemdeclare{misc}{C3}
\bibitem{C3}
\bibinfo{author}{Merlin \surnamestart Carl\surnameend} (\bibinfo{year}{2020}):
  \emph{\bibinfo{title}{Using Automated Theorem Provers for Mistake Diagnosis
  in the Didactics of Mathematics}}, \doi{10.48550/arXiv.2002.05083}.
\newblock \bibinfo{note}{ArXiv:2002.05083v1}.

\bibitemdeclare{misc}{C1}
\bibitem{C1}
\bibinfo{author}{Merlin \surnamestart Carl\surnameend} (\bibinfo{year}{2023}):
  \emph{\bibinfo{title}{Improving the Diproche CNL through Autoformalization
  via Large Language Models}}.
\newblock \eprint{arXiv:2303.17513}.

\bibitemdeclare{inproceedings}{CK}
\bibitem{CK}
\bibinfo{author}{Merlin \surnamestart Carl\surnameend} \&
  \bibinfo{author}{Regula \surnamestart Krapf\surnameend}
  (\bibinfo{year}{2020}): \emph{\bibinfo{title}{Diproche - ein automatisierter
  Tutor f\"ur den Einstieg ins Beweisen}}.
\newblock In: {\slshape \bibinfo{booktitle}{Digitale Kompetenzen und
  Curriculare Konsequenzen}}, pp. \bibinfo{pages}{43--56}.

\bibitemdeclare{phdthesis}{Cr}
\bibitem{Cr}
\bibinfo{author}{Marcos \surnamestart Cramer\surnameend}
  (\bibinfo{year}{2013}): \emph{\bibinfo{title}{Proof-checking mathematical
  texts in controlled natural language}}.
\newblock Ph.D. thesis, \bibinfo{school}{Rheinische
  Friedrich-Wilhelms-Universit\"at Bonn}.

\bibitemdeclare{inproceedings}{GGNS}
\bibitem{GGNS}
\bibinfo{author}{Salwa~Tabet \surnamestart Gonzalez\surnameend},
  \bibinfo{author}{St{\'{e}}phane \surnamestart Graham{-}Lengrand\surnameend},
  \bibinfo{author}{Julien \surnamestart Narboux\surnameend} \&
  \bibinfo{author}{Natarajan \surnamestart Shankar\surnameend}
  (\bibinfo{year}{2021}): \emph{\bibinfo{title}{Semantic parsing of geometry
  statements using supervised machine learning on synthetic data}}.
\newblock In \bibinfo{editor}{Jasmin \surnamestart Blanchette\surnameend},
  \bibinfo{editor}{James~H. \surnamestart Davenport\surnameend},
  \bibinfo{editor}{Peter \surnamestart Koepke\surnameend},
  \bibinfo{editor}{Michael \surnamestart Kohlhase\surnameend},
  \bibinfo{editor}{Andrea \surnamestart Kohlhase\surnameend},
  \bibinfo{editor}{Adam \surnamestart Naumowicz\surnameend},
  \bibinfo{editor}{Dennis \surnamestart M{\"{u}}ller\surnameend},
  \bibinfo{editor}{Yasmine \surnamestart Sharoda\surnameend} \&
  \bibinfo{editor}{Claudio~Sacerdoti \surnamestart Coen\surnameend}, editors:
  {\slshape \bibinfo{booktitle}{Joint Proceedings of the FMM, FVPS,
  MathUI,NatFoM, and OpenMath Workshops, Doctoral Program, and Work in Progress
  at the Conference on Intelligent Computer Mathematics 2021 co-located with
  the 14th Conference on Intelligent Computer Mathematics {(CICM} 2021),
  Virtual Event, Timisoara, Romania, July 26 - 31, 2021}}, {\slshape
  \bibinfo{series}{{CEUR} Workshop Proceedings}} \bibinfo{volume}{3377},
  \bibinfo{publisher}{CEUR-WS.org}.
\newblock \urlprefix\url{https://ceur-ws.org/Vol-3377/natfom5.pdf}.

\bibitemdeclare{book}{H}
\bibitem{H}
\bibinfo{author}{Leon \surnamestart Henkin\surnameend} (\bibinfo{year}{1967}):
  \emph{\bibinfo{title}{Logical Systems Containing Only a Finite Number of
  Symbols}}.
\newblock \bibinfo{publisher}{Presses de l'Universite de Montreal},
  \bibinfo{address}{Montreal,}.

\bibitemdeclare{book}{LN}
\bibitem{LN}
\bibinfo{author}{George \surnamestart Lakoff\surnameend} \&
  \bibinfo{author}{Rafel~E. \surnamestart N{\'u}{\~n}ez\surnameend}
  (\bibinfo{year}{2001}): \emph{\bibinfo{title}{Where mathematics comes from :
  how the embodied mind brings mathematics into being}}.
\newblock \bibinfo{publisher}{Basic Books}.

\bibitemdeclare{inproceedings}{BM}
\bibitem{BM}
\bibinfo{author}{A.~\surnamestart Moreno\surnameend} \&
  \bibinfo{author}{N.~\surnamestart Budesca\surnameend} (\bibinfo{year}{2000}):
  \emph{\bibinfo{title}{Mathematical Logic Tutor-Propositional Calculus}}.
\newblock In: {\slshape \bibinfo{booktitle}{First International Congress on
  Tools for Teaching Logic}}, pp. \bibinfo{pages}{99--106}.

\bibitemdeclare{inproceedings}{WBKU}
\bibitem{WBKU}
\bibinfo{author}{Wang \surnamestart Qingxiang\surnameend},
  \bibinfo{author}{Chad \surnamestart Brown\surnameend},
  \bibinfo{author}{Cezary \surnamestart Kaliszyk\surnameend} \&
  \bibinfo{author}{Josef \surnamestart Urban\surnameend}
  (\bibinfo{year}{2019}): \emph{\bibinfo{title}{Exploration of Neural Machine
  Translation in Autoformalization of Mathematics in Mizar}}.
\newblock In: {\slshape \bibinfo{booktitle}{CPP 2020: Proceedings of the 9th
  ACM SIGPLAN International Conference on Certified Programs and Proofs}}, pp.
  \bibinfo{pages}{85--98}, \doi{10.1145/3372885.3373827}.

\bibitemdeclare{inproceedings}{VLP}
\bibitem{VLP}
\bibinfo{author}{Konstantin \surnamestart Verchinine\surnameend},
  \bibinfo{author}{Alexander~V. \surnamestart Lyaletski\surnameend} \&
  \bibinfo{author}{Andrei \surnamestart Paskevich\surnameend}
  (\bibinfo{year}{2007}): \emph{\bibinfo{title}{System for Automated Deduction
  (SAD): A Tool for Proof Verification}}.
\newblock In: {\slshape \bibinfo{booktitle}{Automated Deduction – CADE-21.
  CADE 2007.}}, \bibinfo{volume}{4603}, \bibinfo{publisher}{Springer, Berlin,
  Heidelberg.}, \doi{10.1007/978-3-540-73595-3\_29}.
\newblock \urlprefix\url{https://api.semanticscholar.org/CorpusID:6915907}.

\bibitemdeclare{inproceedings}{WJLRSJS}
\bibitem{WJLRSJS}
\bibinfo{author}{Yuhuai \surnamestart Wu\surnameend}, \bibinfo{author}{Albert
  \surnamestart Jiang\surnameend}, \bibinfo{author}{Wenda \surnamestart
  Li\surnameend}, \bibinfo{author}{Markus \surnamestart Rabe\surnameend},
  \bibinfo{author}{Charles \surnamestart Staats\surnameend},
  \bibinfo{author}{Mateja \surnamestart Jamnik\surnameend} \&
  \bibinfo{author}{Christian \surnamestart Szegedy\surnameend}
  (\bibinfo{year}{2022}): \emph{\bibinfo{title}{Autoformalization with Large
  Language Models}}.
\newblock In: {\slshape \bibinfo{booktitle}{36th Conference on Neural
  Information Processing Systems (NeurIPS)}}, \doi{10.48550/arXiv.2205.12615}.

\end{thebibliography}

\vfill

\newpage

\section{Appendix}

In this appendix, we give the detailed results of testing the performance of GPT-4-Turbo and WizardCoder-34B in autoformalization of natural language sentence in propositional logic and of GPT-4-Turbo in first-order logic. Although text-davinci-003 was used in the version of the system described above, it is now considered legacy and its use is deprecated; moreover, its performance in autoformalization tasks appears to have considerably dropped since spring 2023; therefore, no results for text-davinci-003 are listed. We first give the prompt given to the LLM, followed by a table listing the input sentences, the model's output and an evaluation of correctness. It should be noted that the sentences were tested in the order in which they are listed in the table and that other orders may lead to different results.

\subsection{Propositional Logic}

For propositional logic, we used the following prompt; in WizardCoder-34B, temperature was set to $0$, repitition penalty to $1$ (the minimum). Below, $\veebar$ is used for exclusive disjunction. In the case of missing brackets, formulas were read according to the usual priority rules. In (31) and (56), a missing closing bracket was added in the WizardCoder output, which was still counted as correct, as this can easily be done automatically.

\begin{tcolorbox}

Express the sentence as a formula in propositional logic, using the given notation. 

Notation:

\begin{itemize}
\item S:``The sun shines''
\item R:``It rains''
\item P:``There will be a party''
\item M:``It is Monday''
\item A:``Andreas will attend the party''
\item B:``Barbara will attend the party''
\item C:``The moon is made of green cheese''
\item G:``There is a giraffe on the moon''
\end{itemize}
\end{tcolorbox}

For the GPT-4-Turbo assistant, the following was added:

\begin{tcolorbox}
    If the given sentence cannot be expressed with the given notation, return ``not expressable''.
\end{tcolorbox}

\begin{small}
\begin{longtable}{|c|p{9em}|c|c|c|c|}
  \hline
  Nr & \multicolumn{1}{c|}{Input} & GPT-4 & +/- & WizardCoder-34B & +/- \\
  \hline 
  \hline
  1&It does not rain and the sun shines. & $\neg R\wedge S$ & + & example &  \\
  \hline
  2&It's neither sunny nor rainy. & $\neg S\wedge \neg R$ & + & $\neg(S\vee R)$ & + \\
  \hline
  3& If there is a party on Monday, Andreas will attend it. & $(P\wedge M)\rightarrow A$ & + & $M\rightarrow A$ & - \\
  \hline
  4& If it rains on Monday, Andreas will stay away from the party. & $(R\wedge M)\rightarrow\neg A$ & + & $(M\wedge R)\rightarrow\neg A)$ & + \\
  \hline
  5& Andreas and Barbara will only attend the party together. & $(A\leftrightarrow B)$ & + & $(A\leftrightarrow B)$ & + \\
  \hline 
%
6&There will be a party if the sun shines on Monday. & $(S\wedge M)\rightarrow P$ & + & $(S\rightarrow P)$ & - \\
\hline
7& It is not true that Andreas will abstain from the party when it rains. & $\neg(R\rightarrow\neg A)$ & + & $\neg((R\rightarrow\neg A)\wedge (M\wedge R))$& - \\
\hline
8&On rainy Mondays, Barbara does not go to parties. & $(R\wedge M)\rightarrow\neg B$ & + &$((M\wedge R)\rightarrow\neg B)$ & + \\
\hline
9&It is a sunny, rainy Monday. & $S\wedge R\wedge M$ &+& $(S\wedge R\wedge M$) & +\\
\hline
10&On sunny Mondays, it rains. & $(S\wedge M)\rightarrow R$ & + & $(M\wedge S)\rightarrow R)$ & + \\
\hline 
11&Andreas and Barbara attend the party. & $A\wedge B$ &+ & $(A\wedge B)$ & +\\
\hline
12&At least one of Andreas and Barbara attends the party.& $A\vee B$ & + & $(A\vee B)$ & +\\
\hline
13&Exactly one of Andreas and Barbara attends the party.& $(A\wedge\neg B)\vee(\neg A\wedge B)$ & + & $(A\veebar B)$& +\\
\hline
14 & Andreas parties provided Barbara doesn't. & $\neg B\rightarrow A$ & + & $(\rightarrow\neg A)\rightarrow P$& - \\
\hline
15 & Andreas goes to the party provided Barbara doesn't. & $\neg B\rightarrow A$ & + & $(B\rightarrow\neg A)$ & +\\
\hline
16&Barbara and Andreas only visit Monday parties.&$(B\vee A)\rightarrow M$ & + &\makecell{$(M\rightarrow (A\vee B))\wedge(\neg A\rightarrow M)$$\wedge(\neg B\rightarrow M)$}& -\\
\hline
17&On Monday, there is good or bad weather.&$M\rightarrow(S\vee R)$ & + &$(M\rightarrow(S\vee R))$&+\\
\hline
18&When it is wet, Andreas does not go to the party, but Barbara does.&$R\rightarrow(\neg A\wedge B)$ & + &$((R\rightarrow\neg B)\wedge B)$& -\\
\hline
19 &It is a bright Monday, and Andreas and Barbara jointly visit the celebration. & $(S\wedge M)\wedge (A\wedge B)$ & + &$(M\wedge S\wedge (A\vee B))$&-\\
\hline
20 & On Tuesday, there is a thunderstorm. & not expressable & + & not expressable & + \\
\hline
21 & Christa exercises on Tuesday. & not expressable & + &not expressable& + \\
\hline
22 & It's a bright Monday. & $S\wedge M$ & + & $(M\wedge S)$& + \\
\hline
23 & It rains if and only if the sun shines. & $R\leftrightarrow S$ & + &$(S\rightarrow R)\wedge(R\rightarrow S)$& +\\
\hline
24 & Sunshine and rain do not occur together & $\neg(S\wedge R)$ & + &not expressible&-\\
\hline
25 & Monday is no day for parties. & $M\rightarrow\neg P$ & + &$\neg M$&- \\
\hline
26 & Andreas is not a party person. & $\neg A$ / not expressable & + &$\neg A$& + \\
\hline
27 & Andreas is not a Tyrannosaurus. & not expressable & + & not expressable & + \\
\hline
28 & Andreas stays away from the party. & $\neg A$ & + &$\neg A$& +\\
\hline
29 & Andreas does not stay away from the party. & $A$ & + &$A$& +\\
\hline
30 &If Andreas stays away from the party, then so does Barbara. & $\neg A\rightarrow\neg B$ & + & $(\neg A\rightarrow \neg B)$ & + \\
\hline
31 & On sunny days, Barbara and Andreas celebrate. & $S\rightarrow(A\wedge B)$ & + & $(S\rightarrow(A\wedge B)$ & + \\ 
\hline
32 & Barbara feasts in the rain, but Andreas doesn't. & $R\rightarrow(B\wedge\neg A)$ & + &$(R\rightarrow (B\wedge \neg A))$& + \\
\hline
33 & It's not Monday, but Barbara still parties. & $\neg M\wedge B$ & + &$\neg M$& - \\
\hline
34 & In spite of the downpour, Andreas celebrates. & $R\wedge A$ & + &$(R\rightarrow A)$& - \\
\hline
35 & It's sun or rain, but not both. & $(S\vee R)\wedge\neg(S\wedge R)$ & + &$(S\veebar R)$& + \\
\hline
36&On sunny and also on rainy days, no parties take place. & $(S\rightarrow\neg P)\wedge (R\rightarrow\neg P)$ & + &not expressible& -\\
\hline
37&On sunny and also on rainy days, there are no parties. & $(S\rightarrow\neg P)\wedge (R\rightarrow\neg P)$ & + &$((S\veebar R)\rightarrow\neg P)$& - \\
\hline
38&It is wrong that the party has not been canceled. & $\neg\neg P$ / $P$ & - &$\neg\neg P$& - \\
\hline
39&Mondays are rainy. & $M\rightarrow R$ & + & $M\wedge R$ & - \\
\hline
40&Rainy days are Andreas' party days. & $R\rightarrow A$ & + & $(R\rightarrow A)$ & + \\
\hline
41&It is a lie that Andreas and Barbara attended the party together. & $\neg(A\wedge B)$ & + &$\neg (A\wedge B)$& + \\
\hline
42&Barbara parties on rainy Mondays. & $(R\wedge M)\rightarrow B$ & + &$(R\wedge M)\rightarrow B$& +\\
\hline
43&It's Monday and it's not. & $M\wedge\neg M$ & + & $(M\wedge\neg M)$ & +\\
\hline
%
44&Andreas goes to the party if he doesn't. & not expressable & - &$(\neg A\rightarrow A)$ & +\\
\hline
45&Barbara goes to the party if and only if Andreas stays at home. & $B\leftrightarrow\neg A$ & + &$(A\rightarrow B)$&-\\
\hline
46&On party Mondays, the sun shines and it rains. & $(P\wedge M)\rightarrow (S\wedge R)$ & + &$(M\rightarrow (S\wedge R))$&- \\
\hline
47&It's Monday if and only if it's Monday. & $M\leftrightarrow M$ & + &$(M\leftrightarrow M)$& + \\
\hline
48&It is neither Monday nor not Monday. & $\neg M\wedge\neg\neg M$ / $\neg M\wedge M$ & + &not expressible& -\\
\hline
49& There is no party if Andreas isn't there. & $\neg A\rightarrow\neg P$ & + &$(\neg A\rightarrow\neg P)$&+ \\
\hline
50& Neither Andreas nor Barbara attend the party. & $\neg A\wedge\neg B$ & + &$(\neg A\vee\neg B)$& + \\
\hline
51&One of Andreas and Barbara does not attend the party. & $(\neg A\vee\neg B)$ & + &$(\neg A\vee\neg B)$&+ \\
\hline
52&If it rains, then Andreas does not attend Monday parties. & $R\rightarrow(M\rightarrow\neg A)$ & + &$(R\rightarrow (M\wedge\neg A))$&+\\
\hline
53& Given that it rains, the sun shines. & $R\rightarrow S$ & + &$(R\rightarrow S)$&+ \\
 \hline
54 &If the moon is made of green cheese, then there is a giraffe on it. & $C\rightarrow G$ & + & $(C\rightarrow G)$ &+\\
\hline
55&There are no moon giraffes. & $\neg G$ & + &$\neg G$& +\\
\hline
56&On Mondays, the moon is made of green cheddar and populated by giraffes. & $M\rightarrow(C\wedge G)$ & + &$(M\rightarrow (C\wedge G))$& + \\
\hline
57&There is an elephant on the moon. & not expressable & + &not expressible& +\\
\hline 
\hline
&mistakes (total)  & 2 & & 17 \\
\hline 
\end{longtable}
\end{small}

\subsection{First-Order Logic}

For first-order logic, WizardCoder-34B showed no satisfying performance; for GPT-4-Turbo, we used the following prompt:

\begin{tcolorbox}
    Express the given sentence as a formula in first-order logic, using the following notation:

\begin{itemize}
\item $D(x)$:$x$ is a dog
\item $B(b)$:$b$ barks
\item $S(a)$:$a$ bites
\item $L(a,b)$:$a$ is larger than $b$
\item fr:Fritz
\item he:Hector
\end{itemize}

If the given sentence cannot be expressed with the given notation, return ``not expressible''.
No comment or explanation; only return the formula.
\end{tcolorbox}

\begin{small}
\begin{longtable}{|c|p{14.5em}|c|c|}
\hline
Nr & \multicolumn{1}{c|}{Input} & Output GPT-4 & +/-? \\
\hline
1 & Fritz is a dog. & $D(fr)$ & + \\
\hline
2 & Hector is taller than Fritz. & $L(he,fr)$ & + \\
\hline
3 & If Hector barks, then he bites. & $B(he)\rightarrow S(he)$ & + \\
\hline
4 & There is a dog. & $\exists{x}D(x)$ & + \\
\hline
5 & Every dog barks. & $\forall{x}(D(x)\rightarrow B(x))$ & + \\
\hline
6 & Some dogs bite. & $\exists{x}(D(x)\wedge S(x))$ & + \\
\hline
7 & Hector does not bite. & $\neg S(he)$ & + \\
\hline
8 & Barking dogs bite. & $\forall{x}((D(x)\wedge B(x))\rightarrow S(x))$ & + \\
\hline
9 & Barking dogs don't bite. & $\forall{x}((D(x)\wedge B(x))\rightarrow\neg S(x))$ & + \\
\hline
10&All dogs bite. & $\forall{x}(D(x)\rightarrow S(x))$ & + \\
\hline
11&Everything bites. & $\forall{x}S(x)$ & + \\
\hline
12&Barking dogs don't bark.& $\forall{x}((D(x)\wedge B(x))\rightarrow\neg B(x))$ & + \\
\hline
13&If Hector barks, he is not a real dog. & not expressible & - \\
\hline
14&If Hector barks, he is not a dog. & $B(he)\rightarrow\neg D(he)$ &+ \\
\hline
15&Fritz and Hector are barking dogs. & $(D(fr)\wedge B(fr))\wedge (D(he)\wedge B(he))$ & + \\
\hline
16&If a dog barks, then Fritz barks. & $\forall{x}((D(x)\wedge B(x))\rightarrow B(fr))$ & + \\
\hline
17&No dog is bigger than Fritz. & $\forall{x}(D(x)\rightarrow\neg L(x,fr))$ & + \\
\hline
18& There is a god. & not expressible. & + \\
\hline
19&Fritz is the tallest dog. & $D(fr)\wedge\neg\exists{x}(D(x)\wedge L(x,fr))$ & + \\
\hline
20&No dog that is larger than Fritz will bite. & $\forall{x}(D(x)\wedge L(x,fr)\rightarrow\neg S(x))$ & +\\
\hline
21&Fritz is smaller than Hector. & $L(he,fr)$ & + \\
\hline
22&One of Hector and Fritz barks. & $B(he)\vee B(fr)$ & + \\
\hline
23& Some dogs are larger than both Fritz and Hector. & $\exists{x}(D(x)\wedge L(x,fr)\wedge L(x,he))$ & + \\
\hline
24&In terms of size, Fritz surpasses himself. & $L(fr,fr)$ & + \\
\hline
25&Some dogs are larger than others. & $\exists{x}\exists{y}(D(x)\wedge D(y)\wedge L(x,y))$ & + \\
\hline
26&Each dog is larger than each dog. & $\forall{x}\forall{y}(D(x)\wedge D(y)\rightarrow L(x,y))$ &+ \\
\hline
27&Fritz is a dog and he isn't. & $D(fr)\wedge\neg D(fr)$ &+ \\
\hline
28&Fritz is the only dog. & $D(fr)\wedge\forall{x}(D(x)\rightarrow x=fr)$ & + \\
\hline
29&Every dog is larger than every other dog. & $\forall{x}\forall{y}((D(x)\wedge D(y)\wedge x\neq y)\rightarrow L(x,y))$ &+ \\
\hline
30&Exactly one of Hector and Fritz barks. & $B(he)\wedge\neg B(fr)\vee\neg B(he)\wedge B(fr)$ &+ \\
\hline
31&Hector is larger than any other dog, except Fritz. & $\forall{x}((D(x)\wedge x\neq fr\wedge x\neq he)\rightarrow L(he,x))$ & + \\
\hline
32& Meowing cats don't scratch. & not expressible & + \\
\hline
33&Biting dogs are not dogs and they don't bite. & $\forall{x}((D(x)\wedge S(x))\rightarrow(\neg D(x)\wedge\neg S(x)))$ & + \\
\hline
34&Hector bites Fritz. & $S(he)$ & - \\
\hline
35&Hector is a baying canine. & not expressible & - \\
\hline
36&Hugo is a biting dog. & not expressible & + \\
\hline
37&If it does not bark, it's not a dog. & $\forall{x}(\neg B(x)\rightarrow \neg D(x))$ & + \\
\hline
38&It's not a dog if it does not bark. & $\forall{x}(\neg B(x)\rightarrow \neg D(x)$ & + \\
\hline
39&There is more than one dog. & $\exists{x}\exists{y}(D(x)\wedge D(y)\wedge x\neq y)$ & + \\
\hline
40&Hector barks, so Fritz does the same. & $B(he)\rightarrow B(fr)$ &+ \\
\hline
41&Fritz barks, and so does Hector. & $B(fr)\wedge B(he)$ & + \\
\hline
42&Barking dogs and also biting dogs are smaller than Fritz. &$\forall{x((D(x)\wedge (B(x)\vee S(x)))\rightarrow L(fr,x))}$ & + \\
\hline
43 & Of each two dogs, one is larger. & $\forall{x}\forall{y}((D(x)\wedge D(y)\wedge x\neq y)\rightarrow (L(x,y)\vee L(y,x)))$ & + \\
\hline
44 & No two dogs are the same. & $\forall{x}\forall{y}((D(x)\wedge D(y)\wedge x\neq y)\rightarrow\neg (x=y))$ & + \\
\hline
45&All dogs are the same: they bite! & $\forall{x}(D(x)\rightarrow S(x))$ & + \\
\hline
46&All dogs are the same. & not expressible & - \\
\hline
47&Hector is a dog that neither barks nor bites. & $D(he)\wedge \neg B(he)\wedge\neg S(he)$ & + \\
\hline
48&Hector and Fritz both bark, but only one of them bites. & $(B(he)\wedge B(fr))\wedge ((S(he)\wedge\neg S(fr))\vee (\neg S(he)\wedge S(fr)))$ & + \\
\hline
49&All dogs bite, but only Fritz and Hector bark. & $\forall{x}(D(x)\rightarrow S(x))\wedge \forall{x}(D(x)\wedge B(x)\rightarrow(x=fr\vee x=he))$ &+  \\
\hline
50&Hector is not the only barker. & $\exists{x}(B(x)\wedge x\neq he)$ & + \\
\hline
\end{longtable}
\end{small}

\end{document}